\documentclass[10pt,twocolumn,letterpaper]{article}

\usepackage{iccv}
\usepackage{times}
\usepackage{epsfig}
\usepackage{graphicx}
\usepackage{amsmath}
\usepackage{amssymb}
\usepackage{multirow}
\usepackage{makecell}

\usepackage{draftwatermark}
\SetWatermarkText{Preprint}
\SetWatermarkLightness{0.925}
\SetWatermarkScale{1}

\usepackage[breaklinks=true,bookmarks=false]{hyperref}

\iccvfinalcopy % *** Uncomment this line for the final submission

\def\iccvPaperID{CVAMD-62} % *** Enter the ICCV Paper ID here

\ificcvfinal\fi

\begin{document}

%%%%%%%%% TITLE
\title{Causality-Driven One-Shot Learning for Prostate Cancer Grading from MRI
\thanks{Preprint accepted on Aug 07, 2023 for ICCV-CVAMD 2023}
}

\author{Gianluca Carloni \dag\textsuperscript{1,2} 
\and Eva Pachetti \dag\textsuperscript{1,2} 
\and Sara Colantonio\textsuperscript{1}\\\\
\textsuperscript{1}Institute of Information Science and Technologies (ISTI-CNR), Italy \\
\textsuperscript{2}University of Pisa, Italy \\
\small \dag These authors contributed equally\\
{\tt\small \{gianluca.carloni, eva.pachetti, sara.colantonio\}@isti.cnr.it}
% For a paper whose authors are all at the same institution,
% omit the following lines up until the closing ``}''.
% Additional authors and addresses can be added with ``\and'',
% just like the second author.
% To save space, use either the email address or home page, not both
}

\maketitle
% Remove page # from the first page of camera-ready.
\ificcvfinal\thispagestyle{empty}\fi

%%%%%%%%% ABSTRACT
\begin{abstract}
  In this paper, we present a novel method to automatically classify medical images that learns and leverages weak causal signals in the image. Our framework consists of a convolutional neural network backbone and a causality-extractor module that extracts cause-effect relationships between feature maps that can inform the model on the appearance of a feature in one place of the image, given the presence of another feature within some other place of the image. To evaluate the effectiveness of our approach in low-data scenarios, we train our causality-driven architecture in a One-shot learning scheme, where we propose a new meta-learning procedure entailing meta-training and meta-testing tasks that are designed using related classes but at different levels of granularity. We conduct binary and multi-class classification experiments on a publicly available dataset of prostate MRI images. To validate the effectiveness of the proposed causality-driven module, we perform an ablation study and conduct qualitative assessments using class activation maps to highlight regions strongly influencing the network's decision-making process. Our findings show that causal relationships among features play a crucial role in enhancing the model's ability to discern relevant information and yielding more reliable and interpretable predictions. This would make it a promising approach for medical image classification tasks.
\end{abstract}

%%%%%%%%% BODY TEXT
\section{Introduction}

Building models that automatically perform diagnoses from medical data could revolutionize a patient's clinical pathway, especially in oncology, minimizing invasive procedures and maximizing the chances of cures for the most severe cases. The implementation of such models in the medical field is severely limited by data availability and the heavy domain shifts that plague the data, which makes the models non-generalizable. Especially in the magnetic resonance imaging (MRI) domain, different vendors yield images with different features, and that prevents the model from being able to generalize well. In a classical fully-supervised manner, the problem can be overcome by training a model with lots of data that covers all the possible data distribution. In practice, however, this is not always possible, due to the limited amounts of labeled/annotated data that affect the medical imaging domain.

Especially when dealing with limited data, one may wish to make an automated recognition model focus on the most discriminative regions of the image instead of paying attention to side details. Traditional convolutional neural networks (CNN) and transformer networks, for instance, can discover features hidden within input data together with their mutual co-occurrence. However, they are weak at discovering and making explicit hidden causalities between the features, which could be the reason behind a particular outcome. Indeed, image classification models are expected to distinguish between the classes of images taking into account the causal relationships between the features from the images, in a way that might resemble how humans accomplish the task. For this reason, several bridges between the field of causality and computer vision are emerging in the literature \cite{scholkopf2021toward, berrevoets2023causal}.

A particular case would be discovering hidden causalities among objects in an image dataset. Unlike tabular data, which have a structured nature, when it comes to images, their representation does not include any explicit indications regarding objects or patterns. Instead, individual pixels convey a particular scene visually, and image datasets do not provide labels describing the objects’ dispositions. Due to this, supervised machine learning as such cannot approach them. Additionally, unlike video frames, from a single image one may not see the dynamics of appearance and change of the objects in the scene. Therefore, a priori information as a hint for causal discovery is absent. 

To approach the problem of learning hidden causalities within images, Lopez-Paz \etal \cite{lopez2017discovering} suggest the “causal disposition” concept as more primitive than interventional causation (do-calculus) and causal graphs from Pearl’s approach \cite{pearl2009causality,pearl2018book}. However, it could be the only way to proceed with limited a priori information. In their view, by counting the number $C(A,B)$ of images in which the causal dispositions of artifacts $A$ and $B$ are such that $B$ disappears if one removes $A$, one can assume that artifact $A$ causes the presence of artifact $B$ when $C(A,B)$ is greater than the converse $C(B,A)$. As a trivial example, imagine the image of a car on a bridge. Now, if we were to remove the car, then this would keep the image realistically looking (i.e., scene consistency), since an observer may see similar scenes among other images. Conversely, if we were to remove the bridge, then this would make the scene inconsistent, as that scenario is likely never seen among other images (i.e., flying cars). Therefore, we may assume that the presence of the bridge has some effect on the presence of the car. This concept leads to the intuition that any causal disposition induces a set of asymmetric causal relationships between the artifacts from an image (features, object categories, etc.) that represent (weak) causality signals regarding the real-world scene. To automatically infer such an asymmetric causal relationship from the statistics observed in an image dataset would be a meeting point with a machine vision system.

In this work, we combine a regular CNN with a causality-extraction module to investigate the features and causal relationships between them extracted during training. We build on ideas from \cite{terziyan2023causality} who suggest a way to compute such an asymmetric measure for possible causal relationships within images, and we propose a new scheme based on feature maps enhancement to enable “causality-driven” CNNs to classify images taking into account hidden causalities within them.
Our hypothesis is that it would be possible and reasonable to get some weak causality signals from the individual images of some medical datasets without adding primary expert knowledge, and leveraging them to better guide the learning phase.
Ultimately, a model trained in such a manner would be able to exploit weak causal dispositions of objects in the image scene to distinguish lesion grades even with limited data and possibly domain shift on the test set. 

To evaluate how these causality-driven networks could behave in a low-data regime, we perform the training in a Few-shot learning (FSL) manner, and in particular in One-shot learning (OSL). Here, we propose a novel training scheme in which we design meta-training and meta-testing tasks having related classes (i.e., the same clinical problem is addressed) but at different granularity levels. To perform such experiments, we exploit the Deep Brownian Distance Covariance (DeepBDC) \cite{Xie22} method.

Our paper is structured as follows. After citing relevant works related to our topic in Sec. \ref{sec:related_works}, we present the rationale behind causality-driven CNNs and our network proposition in Sec. \ref{sec:causality_cnns}, together with a description of the DeepBDC method of our choice in Sec. \ref{sec:one_shot_learning}. Later, in Sec. \ref{sec:experiments}, we dive into the details of our experiments settings, regarding both the dataset used and the meta-training and meta-testing schemes. Finally, in Sec. \ref{sec:results} and Sec. \ref{sec:discussion}, we provide the results of our experiments and discuss our findings, summarizing our key conclusions.

\section{Related works}
\label{sec:related_works}

Several approaches to integrating causality into FSL have been proposed in the literature, leading to different directions and applications. One notable example is the work by Yue \etal \cite{Yue20}, where they leverage causality to demonstrate that pre-training in FSL can act as a confounder, resulting in spurious correlations between samples in the support set and their corresponding labels. To address this issue, they propose a novel FSL paradigm in which they perform causal interventions on the Structural Causal Model (SCM) of many-shot learning employing a backdoor adjustment approach.
Based on that work, Li \etal \cite{Li22} propose a method to mitigate the influence of confounders during the pre-training phase of the prototypical network \cite{Snell17}. They accomplish this by stratifying the pre-trained knowledge using a backdoor adjustment based on causal intervention. Specifically, the backdoor adjustment operations are applied in the metric layer of the prototypical network. The feature vectors of the class prototype and the query set are divided into N equal-size disjoint subsets, and the corresponding subsets are fed into their respective classifiers. The final prediction result is obtained by averaging the prediction probabilities from the N classifiers.
Furthermore, in the work by Yang \etal \cite{Yang23}, the authors propose a method to enhance the robustness and generalizability of few-shot text classification. They achieve this by extracting causal associations from text using a causal representation framework for FSL. The process involves performing causal interventions to generate new data with label-relevant information for each input. The original and augmented texts turned into feature representations are fed into a factorization module, which enforces the separation and joint independence of the representations from non-causal factors. Finally, the feature representations are utilized by a classification module for making predictions.

\section{Methods}
\subsection{Causality-driven CNNs}
\label{sec:causality_cnns}
\paragraph{Preliminaries.}

In automatic image recognition, deep neural network classifiers obtain the essential features required for classification not directly from the pixel representation of the input image but through a series of convolution and pooling operations. These operations are designed to capture meaningful features from the image. Convolution layers are responsible for summarizing the presence of specific features in the image and generating a set of feature maps accordingly. As these maps are sensitive to the spatial location of features in the image, pooling is employed to consolidate the presence of particular features within groups of neighboring pixels in square-shaped sub-regions of the feature map.

\paragraph{Causality signals in images.}
When a feature map $F^i$ contains only non-negative numbers (e.g., thanks to ReLU functions) and is normalized in the interval $[0,1]$, we can interpret its values as probabilities of that feature to be present in a specific location, for instance, $F^i_{r,c}$ is the probability that the feature $i$ is recognized at coordinates ${r,c}$.
By assuming that the last convolutional layer outputs and localizes to some extent the object-like features, we may modify the architecture of a CNN such that the $n \times n$ feature maps ($F^1,F^2,\dots F^k$) obtained from that layer are fed into a new module that computes pairwise conditional probabilities of the feature maps. The resulting $k \times k$ causality map would represent the causality estimates for the features. 

\paragraph{Computing causality maps.}
Given a pair of feature maps $F^i$ and $F^j$ and the formulation that connects conditional probability with joint probability, $P(F^i|F^j) = \frac{P(F^i,F^j)}{P(F^j)}$, following \cite{terziyan2023causality}, we heuristically estimate this quantity regarding the pairs of features by adopting two possible methods, namely \textit{Max} and \textit{Lehmer}.
The \textit{Max} method considers the joint probability to be the maximal presence of both features in the image (each one in its %own 
location):
\begin{equation}
    P(F^i|F^j) = \frac{(\max_{r,c} F^i_{r,c})\cdot (\max_{r,c} F^j_{r,c})}{\sum_{r,c} F^j_{r,c}}
    \label{eq:causality_method_max}
\end{equation}
On the other hand, the \textit{Lehmer} method entails computing 
\begin{equation}
    P(F^i|F^j)_p = \frac{LM_p(F^i \times F^j)}{LM_p(F^j)}
    \label{eq:causality_method_lehmer}
\end{equation}
where $F^i \times F^j$ is a vector of $n^4$ pairwise multiplications between each element of the two $n \times n$ feature maps, while $LM_p$ is the generalized Lehmer mean function \cite{bullen2003handbook} with parameter $p$, which is an alternative to power means for interpolating between minimum and maximum of a vector $x$ via harmonic mean ($p=-2$), geometric mean ($p=-1$), arithmetic mean ($p=0$), and contraharmonic mean ($p=1$):
$LM_p(x) = \frac{\sum_{k=1}^n x_k^p}{\sum_{k=1}^n x_k^{p-1}}$.
Equations \ref{eq:causality_method_max} and \ref{eq:causality_method_lehmer} could be used to estimate asymmetric causal relationships between features $F^i$ and $F^j$, since, in general, $P(F^i|F^j) \neq P(F^j|F^i)$. By computing these quantities for every pair $i$ and $j$ of the $k$ feature maps, the $k \times k$ causality map is obtained. We interpret asymmetries in such probability estimates as weak causality signals between features, as they provide some information on the cause-effect of the appearance of a feature in one place of the image, given the presence of another feature within some other places of the image. Accordingly, a feature may be deemed to be the reason for another feature when $P(F^i|F^j) > P(F^j|F^i)$, that is ($F^i \rightarrow F^j$), and vice versa.  

\paragraph{Embedding causality in regular CNNs.}
Once the causality map is computed, it can be embedded into the basic CNN architecture. Terziyan and Vitko \cite{terziyan2023causality} flatten these suggested causality estimates, concatenate them to the set of flattened feature maps, and let the CNN learn how these estimates might influence image classification. Differently from them, we exploit the causality map in a new way and get a weighting vector to enhance or penalize the single feature maps during training.

\paragraph{Causality weights.}
At each epoch, as training progresses, we look for asymmetries between elements opposite the main diagonal of the causality map. Some features may be more often found on the left side of the arrow (i.e., $F\rightarrow$) than on the right side (i.e., $\rightarrow F$). Therefore, we use such learned causalities to compute causality weights to assign a degree of importance to each feature map. Specifically, for each feature map, we take as its weighting factor the difference between the number of times it was found to cause other feature maps and the number of times it was found to be caused by another feature map. Computing such quantity for every feature results in a vector of causality factors, which is then passed through a ReLU activation to set to zero all the negative elements.

\paragraph{Models.}
We propose two variants of the model:
\begin{itemize}
    \item \textbf{mulcat} (\textit{multiply and concatenate}). The non-negative causality factors multiply the corresponding feature maps, resulting in a causality-driven version of these feature maps. In this enhanced version, each feature map is strengthened according to its causal influence within the image's scene. Those features are merged with the original features by concatenation along the channel axis and form the final feature set that influences the classification outcomes.
    \item \textbf{mulcatbool}. Same as the previous, but before multiplication, the factors undergo boolean thresholding where all the non-zero factors are assigned a new weight of $1$, while $0$ otherwise.
\end{itemize}
The first method weighs features more according to their causal importance (a feature that is \textit{cause} $10$ times more than another receives $10$ times more weight). In contrast, the second method is more conservative and assigns all features that are most often \textit{causes} the same weight. We experiment with both of them and compare their results.

\subsection{One-shot learning}
\label{sec:one_shot_learning}

In standard FSL, the training process occurs in episodes or tasks. Each task is formulated as an \emph{$N$-way $K$-shot} classification problem, where $N$ represents the number of classes, and $K$ is the number of \textit{support} images per class. We refer to $Q$ as the number of \textit{query} images per class. For our experiments, we specifically focused on \emph{N-way 1-shot} classification, and we employed the DeepBDC method introduced by Xie \etal \cite{Xie22}. In particular, we utilized the meta-learning implementation of DeepBDC, known as Meta DeepBDC. DeepBDC is a metric-based FSL method that employs the BDC as the distance measure between prototypes. The BDC is defined as the Euclidean distance between the joint characteristic function and the product of the marginal of two random variables $X \in \mathbb{R}^p$ and $Y \in \mathbb{R}^q$. Following \cite{Xie22}, we provide a more formal definition of BDC:

\begin{equation}
    \rho(X,Y) = \int_{\mathbb{R}^p}^{}\int_{\mathbb{R}^q}^{}\frac{|\Phi_{XY}(t,s)-\Phi_X(t)\Phi_Y(s)|^2}{c_pc_q||t||^{1+p}||s||^{1+q}}dtds,
\end{equation}

\noindent where $\Phi_{X}(t)$ and $\Phi_{Y}(s)$ are the marginal distributions of X and Y, respectively, $\Phi_{XY}(t,s)$ is the joint characteristic function of the two random variables and $c_p$ is defined as $c_p = \pi^{(1+p)/2}/\Gamma((1+p)/2)$, where $\Gamma$ is the complete gamma function.  
DeepBDC has demonstrated higher performance compared to state-of-the-art methods while being straightforward to deploy since it can be implemented as a parameter-free spatial pooling layer that accepts feature maps as input and provides a BDC matrix. 

\section{Experiments}
\label{sec:experiments}
\subsection{Dataset and pre-processing}

In our study, we conducted meta-training, meta-validation, and meta-testing using the publicly available $1500$-acquisition dataset from the PI-CAI challenge \cite{Saha23}. This dataset comprises mpMRI acquisitions of the prostate, and for our experiments, we focused exclusively on cancerous patients. In particular, we selected only T2-weighted (T2w) images containing lesions by exploiting the expert annotations provided in the dataset.
The dataset contained biopsy reports expressing the severity of each lesion as Gleason Score (GS). The pathologists assign a score of $1$ to $5$ to the two most common patterns in the biopsy specimen based on the tumor severity. The two grades are then added together to determine the GS, which can assume all the combinations of scores from "$1$+$1$" to "$5$+$5$". Additionally, the dataset included the assigned GS's group affiliation, defined by the International Society of Urological Pathology (ISUP) \cite{Egevad16}, ranging from $1$ to $5$, which provides the tumor severity information at a higher granularity level. From an even more high-level perspective, lesions with a GS $\leq 3+3$ (ISUP = $1$) and with GS $= 3+4$ (ISUP = $2$) are considered low-grade (LG) tumors, as patients with such lesions typically undergo active surveillance \cite{Mohler19}. Conversely, lesions with GS $> 3+4$ (ISUP $> 2$) are high-grade (HG) tumors, as treatment is foreseen \cite{Mohler19}. In this study, we considered only lesions whose GS was $\geq 3+4$ (ISUP $\geq 2$), as lesion annotations were not provided for ISUP-$1$ lesions. As a result, we had eight classes of GS and four classes of ISUP in our dataset.
The total number of images we used was $2049$ (from $382$ patients), which we divided into training, validation, and testing subsets. Specifically, we used $1611$ images for training, $200$ for validation, and $238$ for testing. During the splitting process, we ensured patient stratification, i.e., all the images of the same patient were grouped in the same subset, avoiding any data leakage.
To replicate a realistic scenario involving distinct distributions in training and testing data, we utilized data from two different vendors: SIEMENS vendor data for meta-training and Philips vendor data for both meta-validation and meta-testing. Indeed, we chose the same validation and test distributions since, as highlighted by Setlur \etal \cite{Setlur21}, using validation samples that are not independent and identically distributed with the test samples can lead to unreliable results when determining the optimal model, specifically the one that maximizes performance on the test set.

As for the data pre-processing, we utilized the provided whole prostate segmentation to extract the mask centroid for each slice. We standardized the field of view (FOV) at $100$ mm in both $x$ ($FOV_x$) and $y$ ($FOV_y$)  directions to ensure consistency across all acquisitions and subsequently cropped each image based on this value around the found centroid. To determine the number of rows ($N_{rows}$) and columns ($N_{cols}$) corresponding to the fixed FOV, we utilized the pixel spacing in millimeters along the x-axis (denoted as $px$) and the y-axis (denoted as $py$). The relationships used to derive the number of columns and rows are $N_{cols} = \frac{FOV_x}{px}$ and $N_{rows} = \frac{FOV_y}{py}$, respectively.
Additionally, we resized all images to a uniform matrix size of $128 \times 128$ pixels to maintain a consistent pixel count. Finally, we performed image normalization using an in-volume method. This involved calculating the mean and standard deviation (SD) of all pixels within the volume acquisition and normalizing each image based on these values using a z-score normalization technique.

\subsection{Classification experiments}

In our study, we conducted experiments on two classification scenarios: (i) distinguishing between LG and HG lesions and (ii) ISUP grading. For each scenario, we carefully designed meta-training and meta-testing tasks.
Most FSL approaches typically involve using unrelated classes between meta-training and meta-testing tasks \cite{Dai23,Mahapatra22,Paul21}. Instead, we propose a different approach where we focus on the same clinical problem but at varying levels of granularity. Specifically, during meta-training, we designed more challenging tasks, requiring the model to distinguish between classes with higher levels of granularity. Conversely, during meta-testing, we provided the model with easier classification tasks involving higher-level classification. Our rationale is that this approach would lead to higher performance on the specific task of interest performed during meta-testing, as the model would find it relatively easier to execute due to its exposure to more complex tasks during meta-training. Below we provide a detailed explanation of how we designed our meta-training and meta-testing tasks for both experiments.

In the first scenario, we labeled the meta-training data according to the four ISUP classes. The model performed binary classification in each meta-training task between two randomly selected classes from the four provided. However, during meta-testing, the model was tasked with a higher-level classification, namely, distinguishing between LG and HG lesions. For ease of reference, we will refer to this experiment as the \emph{2-way} experiment. In the second scenario, we labeled the meta-training data based on the GS. Each training task required the model to distinguish between four randomly selected GS classes out of the total eight. For meta-validation and meta-testing, we labeled each patient based on the ISUP tumor severity score and made the model distinguish across these four classes. Henceforth, we will refer to this experiment as \emph{4-way} experiment. We summarized the labeling procedure for the two experiments in Table \ref{tab:data_labeling}.
\begin{table*}
    \centering
    \begin{tabular}{c|c|c}
         \textbf{Experiment} & \textbf{Splitting} &\textbf{Labels}\\
         \hline

         \multirow{3}{*}{2-way} & Meta-training & \makecell{ISUP $2$, ISUP $3$, ISUP $4$, ISUP $5$}\\
         \cline{2-3}
         & Meta-validation & \textbf{LG} (ISUP $2$) - \textbf{HG} (ISUP $3$, ISUP $4$, ISUP $5$)\\
         \cline{2-3}
        & Meta-test &  \textbf{LG} (ISUP $2$) - \textbf{HG} (ISUP $3$, ISUP $4$, ISUP $5$)\\
        
         \hline
         
         \multirow{3}{*}{4-way} & Meta-training & \makecell{GS $3+4$, GS $4+3$, GS $4+4$, GS $3+5$, GS $5+3$, GS $4+5$, GS $5+4$, GS $5+5$}\\
         \cline{2-3}

         & Meta-validation & ISUP $2$, ISUP $3$, ISUP $4$, ISUP $5$\\
         \cline{2-3}
        & Meta-test & ISUP $2$, ISUP $3$, ISUP $4$, ISUP $5$\\
       
    \end{tabular}
    \caption{A summary of the labeling procedure according to our training approach. ISUP = International Society of Urological Pathology, LG = Low Grade, HG = High Grade, GS = Gleason Score.}
    \label{tab:data_labeling}
\end{table*}

In both scenarios, we employed a one-shot setting for both meta-training and meta-testing. This means that the model only observed a single example per class in the support set of each task. However, during the evaluation phase, we expanded the query set by utilizing ten samples per class in both meta-training and meta-testing tasks.

\subsection{Architecture and training}

Many widely used architectures for large-scale image recognition incorporate an adaptive average pooling layer with an output size of $1\times1$ placed just before the classifier. Its primary advantage is the ability to accommodate input images of varying sizes, as it automatically adjusts its parameters (such as kernel size, stride, and padding) to ensure that the output is consistently $1\times1$ in shape. This dimensionality reduction, however, conflicts with the 2D nature of feature maps for computing causalities. Therefore, since we chose the ResNet18 as the backbone architecture in our work, we substituted its \textit{AdaptiveAvgPool2D} layer with an identity layer in our experiments.
We performed an optimization over the method by which computing the causality maps (i.e., \textit{Max} or \textit{Lehmer}) and, for the \textit{Lehmer} case, over six different values of its parameter \textit{p}: [$-100, -2, -1, 0, 1, 100$]. Accordingly, we trained seven models for each causality setting (i.e., \textit{mulcat} or \textit{mulcatbool}), resulting in $14$ causality-driven models plus one baseline model for each experiment (i.e., 2-way and 4-way). For each of the two causality settings, we chose the best-performing model on the meta-validation set.

\begin{figure*}
\begin{center}
   \includegraphics[width=0.9\linewidth]{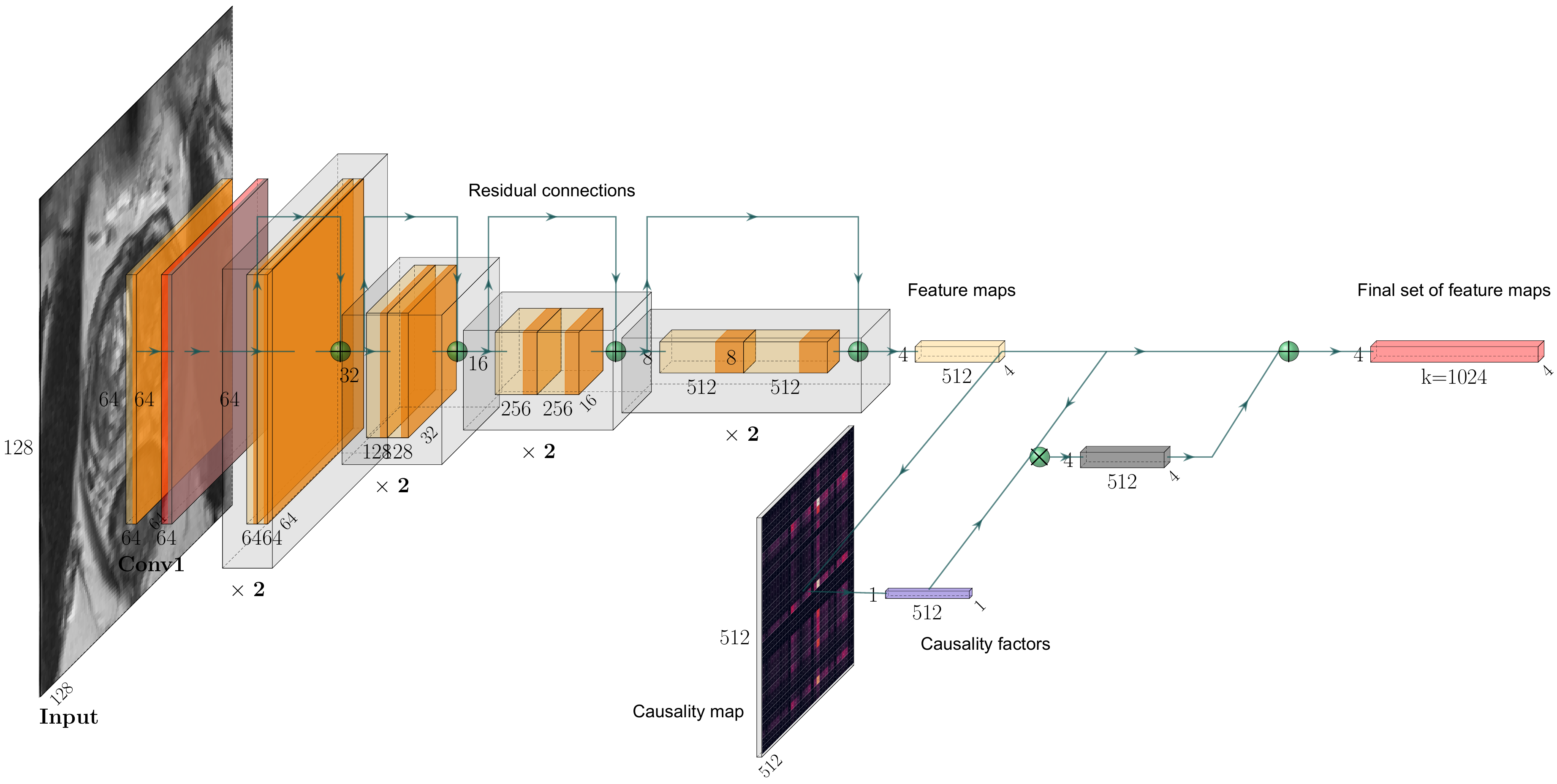}
\end{center}
   \caption{Causality-Driven ResNet18 for prostate cancer Grading from MRI. Here, the \textit{causality map} can be computed with one of \textit{Max} and \textit{Lehmer} options, while the \textit{causality factors} can be computed using either \textit{mulcat} or \textit{mulcatbool} methods. For visualization purposes, the size of the box representing the causality map has been reduced.}
\label{fig:resnet18_causality}
\end{figure*}

Given an input image, the causality-driven ResNet18 extracts $512$ bidimensional feature maps of shape $4\times4$. While those features proceed along the main branch of the network, a copy of them enters the causality module where for each pair, we extract their conditional probabilities by either applying Eq. \ref{eq:causality_method_max} or Eq. \ref{eq:causality_method_lehmer} depending on the causality method of choice. Starting from the resulting  $512\times512$ causality map, the vector of $512$ causality factors is obtained according to the model variant of choice (i.e., \textit{mulcat} or \textit{mulcatbool}) and then multiplied for the corresponding feature maps. Then, after concatenation of two such feature sets, we obtain a set of $1024$ feature maps of shape $4\times4$ for each input image. 
Figure \ref{fig:resnet18_causality} shows the proposed causality-driven network. Although the training is performed task by task, here we represented the functioning of our method for just one input image.

At this point, the final set of feature maps is used to calculate the image representations. Following the Prototypical Networks \cite{Snell17} approach, the classification is performed by computing the BDC between the prototypes of each supported class, calculated as the mean of the BDC matrix of each support image of that class and each query image representation. 
To infuse the model with robustness to different data selections, we performed $600$ meta-training tasks, $600$ meta-validation tasks, and $600$ meta-testing tasks for each experiment.
As our loss function and optimizer, we employed the AUC margin loss (AUCM) \cite{Yang22} and the Proximal epoch stochastic method (PESG) \cite{Guo23}, respectively. These were employed to maximize the Area Under the ROC curve (AUROC), which is more stable w.r.t accuracy to the dataset unbalancing. In addition, we performed our experiments with the following hyperparameters: initial learning rate = $1e-2$, weight decay = $1e-2$, number of epochs = $100$, decay epochs: [$20$,$80$]. At each decay epoch, the learning rate value is divided by $10$.

\subsection{Evaluation metrics}
In our evaluation, we utilized the AUROC as the performance metric for all our experiments. For the $2$-way experiment, we computed the classical binary AUROC by considering the HG as the positive class. In the case of the $4$-way experiment, instead, we calculated the AUROC using the \emph{One-vs-rest} setting. This approach involves computing the AUROC for each class against the rest of the classes independently. In addition, we evaluated the binary classification performance of the models in the $4$-way experiment by computing the AUROC of ISUP class $2$ versus all the rest.

\subsection{Visualizing the impact of causality}
To test the hypothesis that a causally trained model can learn more discriminative representations for image classes, we performed post hoc explainability evaluations. Investigating the variability of visualization results with different choices of post hoc explainable AI (XAI) methods is beyond the scope of our work, therefore, we employed the popular Grad-CAM \cite{selvaraju2017grad}, which belongs to the broad literature on class activation maps (CAM) 
The basic idea behind Grad-CAM is to utilize the gradients flowing back from a chosen layer of a CNN to understand which parts of the image contribute the most to the activation of a particular class. In this case, we chose to compute the Grad-CAM heatmaps at the BDC module level, which takes as input the final set of feature maps. In addition, to make a fair comparison, we computed the heatmaps w.r.t. the ground-truth target of each image and only in the cases for which the prediction was performed correctly by both the non-causality-driven model and the mulcat and mulcatbool models.

\section{Results}

\label{sec:results}

\subsection{Performance of causality-driven CNNs}
The main results of our analysis are reported in Table \ref{tab:best_results}. We reported all those values as mean and SD AUROC across all the $600$ meta-test tasks.
Concerning the $2$-way experiment (i.e., LG vs HG), the baseline model achieved $0.539$ ($0.141$), and embedding the causality module led the model to improve, obtaining $0.550$ ($0.144$) and $0.556$ ($0.141$) AUROC for the \textit{mulcat} and \textit{mulcatbool} variants, respectively. In particular, both these causality-driven variant results were obtained with \textit{Lehmer} causality setting, using a \textit{Lehmer} parameter of -$100$.
Similar behaviour, with more pronounced improvement, was observed for the $4$-way experiment (i.e., ISUP $2-5$), where we obtained $0.585$ ($0.068$) multi-class AUROC for the non-causality-driven model, whereas the \textit{mulcat} and \textit{mulcatbool} variants achieved $0.611$ ($0.069$) and $0.614$ ($0.067$), respectively. Again, both best-performing mulcat and mulcatbool variants were obtained employing the \textit{Lehmer} setting, with \textit{Lehmer} parameters equal to $1$ and -$2$, respectively.

Table \ref{tab:best_results} also shows the results of an ablation study. Indeed, since in the causally-driven implementations is the \textit{causality factors} vector that ultimately determines which (and how) feature maps are enhanced, we modify that vector to weigh features in a random manner rather than based on a principled way based on the causality map. The ablation variants of the \textit{mulcat} and \textit{mulcatbool} models that we realized are:
\begin{itemize}
    \item \textbf{ablation mulcat}. The $1 \times k $ vector of causality factors (i.e., weights) is replaced with a random vector of the same size with integer values ranging from $0$ (a feature map is never \textit{cause} of another feature) and $k-1$ (it is \textit{cause} of every other feature).
    \item \textbf{ablation mulcatbool}. Similar to the previous, the values of the weights are randomly assigned either to $0$ or to $1$.
\end{itemize}

\begin{table*}
    \centering
    % \resizebox{\textwidth}{!}{
    \begin{tabular}{c|c|c|c}
    \hline
      % \multirow{2}{*}{\textbf{Causality setting}} 
      %\cline{2-3}
       \textbf{Setting} & \textbf{2-way 1-shot} & \textbf{4-way 1-shot} & \textbf{4-way 1-shot*}\\ 
         \hline
         \multicolumn{4}{c}{\textbf{Main results}}\\
         \hline
           
         Non causality-driven & 0.539 (0.141) & 0.585 (0.068) & 0.586 (0.118)\\
         \hline
         Causality-driven mulcat & 0.550 (0.144) & 0.611 (0.069) & 0.712 (0.118) \\
         \hline
         Causality-driven mulcatbool & \textbf{0.556} (0.141) & \textbf{0.614} (0.067) & \textbf{0.713} (0.119) \\
        \hline
         \multicolumn{4}{c}{\textbf{Ablation study}}\\
         \hline

         Ablation mulcat & 0.535 (0.143) &   0.557 (0.063) & 0.557 (0.111)\\
         \hline
         Ablation mulcatbool & 0.540 (0.139) &  0.571 (0.068) & 0.612 (0.119)\\ 
         \hline
    \end{tabular}
    % }
    \caption{Results of the best-performing models w.r.t the causality setting and causality method compared to the non-causality choice for 2-way and 4-way experiments. \textbf{*}: trained to distinguish four classes (ISUP $2-5$), but the AUROC is computed between ISUP $2$ vs. rest.}
    \label{tab:best_results}
\end{table*}

\subsection{Visualizing activation maps}
\begin{figure*}
\begin{center}
   \includegraphics[width=0.8\linewidth]{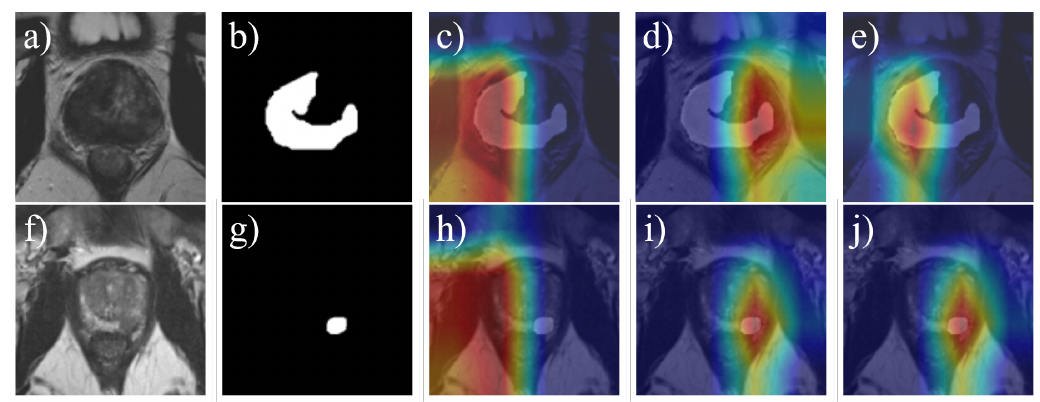}
\end{center}
   \caption{Impact of causality on the learned models. From left to right, the columns represent the input test images, the binary masks for the annotated lesion, the Grad-CAM activations for the baseline models (not causality-driven), the Grad-CAM activations for the \textit{mulcat} models, and the Grad-CAM activations for the \textit{mulcatbool} models. Each row represents a different test case. This image is best seen in colour.}
\label{fig:CAM_causality}
\end{figure*}

As a result of the post hoc explainability evaluations, we obtain the visualizations shown in Figure \ref{fig:CAM_causality}. The first row (\textit{a} to \textit{e}) regards a test case from models trained in the $2$-way setting, while the second row (\textit{f} to \textit{j}) pertains to a case from $4$-way models. From left to right, the columns represent the input test image, the annotation of the lesion in terms of binary masks, the Grad-CAM activation for the baseline models (not causality-driven), the Grad-CAM activation for the \textit{mulcat} models, and the Grad-CAM activation for the \textit{mulcatbool} models.

\section{Discussion and Conclusion}

\label{sec:discussion}
In this study, we investigated the impact of integrating a new causality-extraction module into traditional CNNs to enhance classification performance. We trained this causality-driven model using an OSL approach, leveraging meta-learning conditions with the MetaDeepBDC model \cite{Xie22}. We aimed to assess the effectiveness of such a model in situations where only a few samples are available for training, a challenge frequently encountered in medical image analysis.

In Pearl's terms, our work regards the first rung of the ladder of causation, where reasoning is limited to conditional probabilities based on observational datasets. However, we aimed to explore whether this approach could yield improvements in a scenario involving image data (rather than structured tabular data), and no prior knowledge of the data generation process. Our findings demonstrate that incorporating a causality-driven module into our model leads to enhanced performance compared to the baseline.
This behavior is evident in both the $2$-way and $4$-way experiments. In particular, in the $4$-way experiment, the causality module provided a $3$\% improvement over the baseline in terms of the multi-class AUROC and about $13$\% improvement in terms of ISUP $2$ vs. rest AUROC.

We additionally validated our numerical results both quantitatively and qualitatively.
Quantitatively, we performed ablation studies on the actual impact of the causality factors on producing valuable causality-driven feature maps. As expected, when the causal weights are replaced with random vectors, the accuracy of the final model is worse than its causally-driven counterpart (see Table \ref{tab:best_results}). This seems to suggest that, albeit weak, the causality signals learned during training help the network. 
Qualitatively, we generated Grad-CAM heatmaps to highlight the regions of the input image that strongly influence the network's output. Figure \ref{fig:CAM_causality} presents examples for both the $2$-way and the $4$-way experiments. In both cases, we calculated the heatmaps w.r.t. the ground truth target when all three types of models correctly classified the images.
The heatmaps reveal distinct patterns between the baseline model and the causality-driven models. The former tends to focus on a larger area of the image, including regions outside the prostate and lesion boundaries. Conversely, the causality-driven models concentrate on smaller areas, predominantly encompassing the prostate and the lesion. Figure \ref{fig:CAM_causality} (c-e), which depicts the $2$-way experiment, shows that the baseline model (Figure \ref{fig:CAM_causality} c) primarily attends to the left half of the image, encompassing the lesion as well as non-prostate tissues. In contrast, the \textit{mulcat} version (Figure \ref{fig:CAM_causality} d) exhibits a more focused heatmap, highlighting mainly the prostate and a portion of the lesion. The \textit{mulcatbool} case (Figure \ref{fig:CAM_causality} e) further refines the focus by emphasizing the prostate and a larger portion of the lesion.
Similarly, as for the $4$-way experiment, the baseline model (Figure \ref{fig:CAM_causality} h) pays attention to the left half of the image. In contrast, the \textit{mulcat} and \textit{mulcatbool} versions (Figure \ref{fig:CAM_causality} i-j) prioritize the lesion's immediate surroundings. Although all three models produce accurate predictions, the heatmaps demonstrate that the causality-driven module enables better localization of relevant regions, providing more reliable explanations for the model's predictions.

Comparing the $4$-way experiment to the $2$-way experiment, the former produced better classification results. Indeed, despite being a more complex task, the mean AUROC across all classes is higher. We argue that two factors contribute to this outcome, both associated with the meta-training phase.
Firstly, in the $4$-way experiment, the model encounters a more diverse range of tasks, as the four classes can be selected from a pool of eight distinct options. In contrast, the $2$-way experiment encompasses only four classes in total. Secondly, for the $4$-way experiment, in each meta-training task, the model is trained to distinguish a higher number of classes, representing a more challenging task w.r.t. the binary case. Consequently, the model is better equipped to handle the testing phase, resulting in improved performance. This superiority becomes even more evident when examining the models' performance in the $4$-way experiment in classifying ISUP class $2$ (representing LG lesions) against all other classes (representing HG lesions). Notably, when the \textit{mulcat} and \textit{mulcatbool} causality-driven modules are embedded into the model, the AUROC value for this particular task increases by almost $16$\%. 

Our work comes with several limitations. We only used ResNet18 as our backbone network, and this might have limited the opportunity to find better-suited architectures able to detect finer details in the image and consequently extract more informative latent representations. In addition, we performed our experiments only in an OSL setting, limiting the classification performance of our models. In fact, we must note that the performance values obtained are not yet sufficient for clinical practice. Finally, we validated our method on only one dataset.

Despite that, our findings indicate that integrating a causality-driven module into a classification model can enhance performance, even with severe data limitations, which are common in medical imaging. The causality-driven approach not only improves overall classification results but also helps the model focus more accurately on the critical regions of the image, leading to more reliable and robust predictions. This aspect is particularly critical in medical imaging, where precise and reliable classification is crucial for effective diagnosis and treatment planning.   

{\small
\bibliographystyle{ieee_fullname}
\bibliography{egbib}
}

\end{document}